\pdfoutput=1

\documentclass[11pt]{article}

\usepackage[]{EMNLP2022}

\usepackage{times}
\usepackage{latexsym}

\usepackage[T1]{fontenc}

\usepackage[utf8]{inputenc}

\usepackage{microtype}

\usepackage{inconsolata}

\usepackage{graphicx}
\usepackage{color, soul, multirow}
\usepackage{subfig}
\usepackage{bm}
\usepackage{verbatim}
\usepackage{amssymb}
\usepackage{pifont}
\usepackage{float}
\usepackage{amsmath}
\usepackage{booktabs}
\usepackage{pifont}
\usepackage{tikz}
\usepackage{pgfplots}
\usepackage{lipsum}  
\title{Joint Audio/Text Training for Transformer Rescorer  \\
of Streaming Speech Recognition}
 
\author{Suyoun Kim \hspace{1cm} Ke Li  \hspace{1cm} Lucas Kabela \\
{\bf Rongqing Huang} \hspace{1cm} {\bf Jiedan Zhu} \hspace{1cm} {\bf Ozlem Kalinli} \hspace{1cm} {\bf Duc Le} \\
  Meta, USA  \\
  \texttt{suyounkim@meta.com} \\}

\begin{document}
\maketitle
\begin{abstract}
    Recently, there has been an increasing interest in two-pass streaming end-to-end speech recognition (ASR) that incorporates a 2nd-pass rescoring model on top of the conventional 1st-pass streaming ASR model to improve recognition accuracy while keeping latency low. One of the latest 2nd-pass rescoring model, Transformer Rescorer, takes the n-best initial outputs and audio embeddings from the 1st-pass model, and then choose the best output by re-scoring the n-best initial outputs. However, training this Transformer Rescorer requires expensive paired audio-text training data because the model uses audio embeddings as input. In this work, we present our Joint Audio/Text training method for Transformer Rescorer, to leverage unpaired text-only data which is relatively cheaper than paired audio-text data. We evaluate Transformer Rescorer with our Joint Audio/Text training on Librispeech dataset as well as our large-scale in-house dataset and show that our training method can improve word error rate (WER) significantly compared to standard Transformer Rescorer without requiring any extra model parameters or latency.
\end{abstract}

\section{Introduction}
    Streaming end-to-end automatic speech recognition (ASR) models aim to transcribe the user’s voice with minimal latency and have been widely used in numerous interactive ASR applications that support direct user interaction in real-time. Unlike non-streaming end-to-end ASR \citep{chorowski2015attention, chan2016listen, bahdanau2016end, kim2017joint, Chiu18}, streaming end-to-end ASR, such as RNN-T \citep{Graves12transduction, Prabhavalkar17, battenberg2017exploring, he2019streaming, li2019improving}, are limited to use short audio context or not use future context to satisfy low latency constraints and suffer from higher word error rates (WER). 
    
    \begin{figure}[t]
        \begin{minipage}[b]{1\linewidth}
        \centering
        \centerline{\includegraphics[width=8cm]{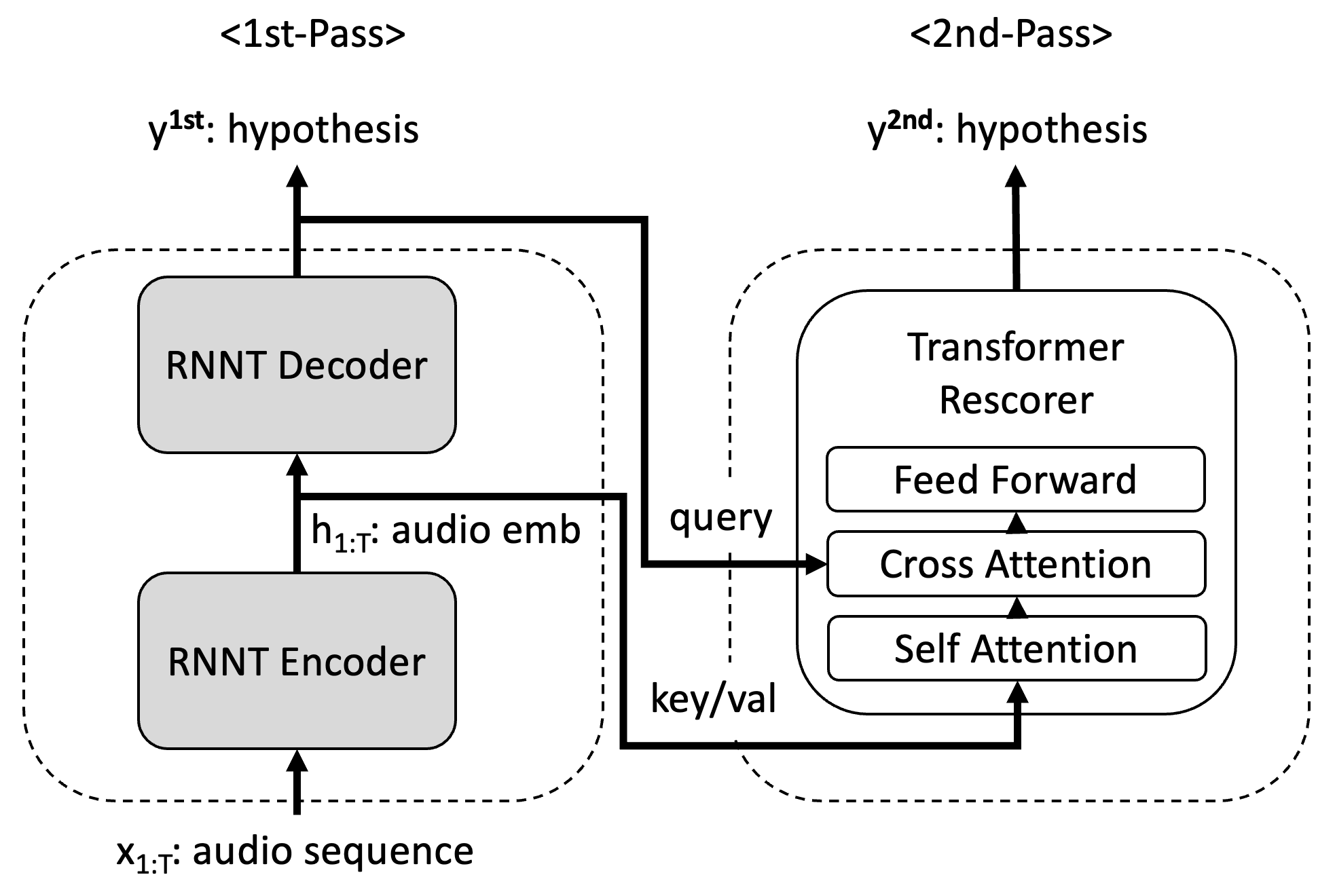}}
        \caption{The two-pass system consists of streaming 1st pass RNN-T model and non-streaming 2nd pass Transformer Rescorer.}
        \label{fig:two_pass}
        \end{minipage}
    \end{figure}
    
    To address this issue of streaming ASR, a Two-Pass architectures has been recently proposed to improve WER while keeping latency low \citep{sainath2019two, li2020parallel, xu2022rescorebert}. The main idea of the Two-Pass architectures is to use a non-streaming model, so-called \textit{Rescorer}, (2nd-pass) on top of the conventional streaming model (1st-pass) to re-score and choose the best output among the initial n-best outputs generated from 1st-pass model. Specifically, the latest Transformer-based \citep{vaswani2017attention} Rescorer \citep{li2020parallel} has been shown promising results in both accuracy and latency improvement. The Rescorer can improve WER and keep low latency because 1) it does not need to perform expensive beam search decoding process which is done already with 1st-pass model, and 2) it exploits full context of audio embeddings generated from the 1st-pass model. While there are performance advantages to exploit full context of audio embeddings as input, training Transformer Rescorer needs expensive paired audio/text data similar to ASR training and thus cannot leverage unpaired text-only data.
    
    \begin{figure*}[t]
        \begin{minipage}[b]{1\linewidth}
        \centering
        \centerline{\includegraphics[height=1.6in]{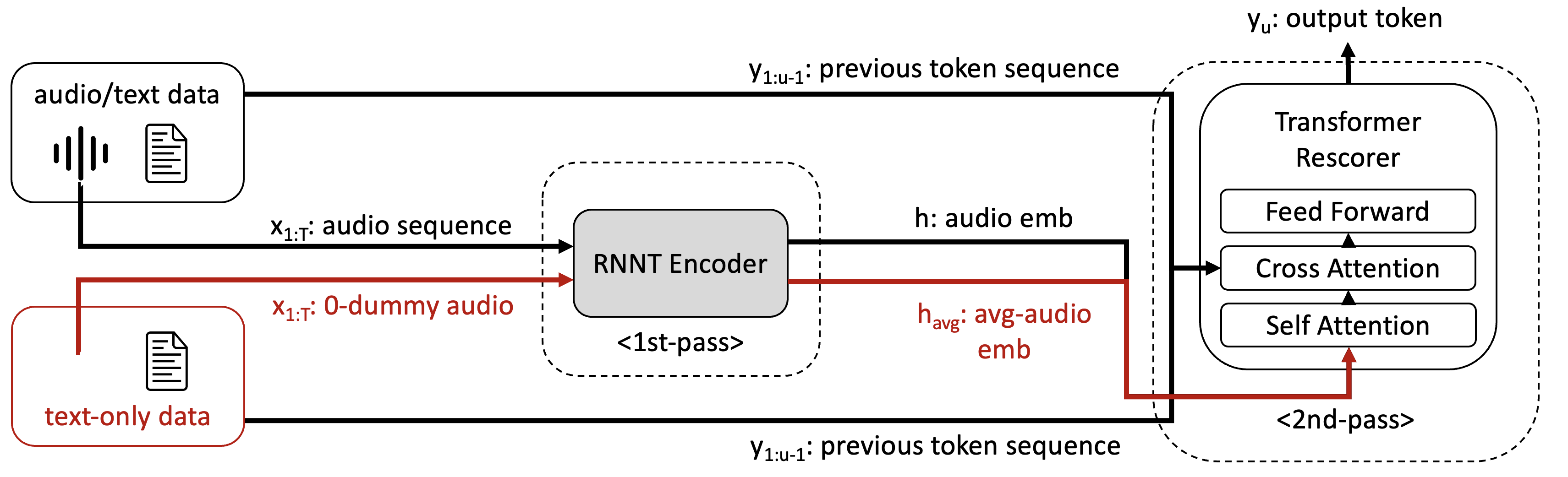}}
        \caption{Our Joint Audio/Text training for Transformer Rescorer in ASR}
        \label{fig:arch}
        \end{minipage}
    \end{figure*}
    
    Many previous studies have investigated approaches to leverage unpaired text-only data including fusion with an external neural network language model (NNLM) \citep{gulcehre2015using, kannan2018analysis, sriram2017cold, shan2019component, toshniwal2018comparison, kannan2018analysis, McDermott19, variani2020hybrid, weinstein2020stateless, kim2021improved}, however, these techniques require extra model parameters and latency. More recently, new approaches to use text-only data without requiring extra model parameters and latency by multi-task learning for LAS \citep{chan2016listen} have been proposed \citep{sainath2020attention, wang2020multitask, tang2022unified}. 
    
    In this work, we present our Joint Audio/Text training method for Transformer Rescorer to leverage unpaired text-only data without requiring any extra parameter or latency. Unlike previous studies \citep{sainath2020attention, wang2020multitask}, our method does not need to use Text-To-Speech (TTS) or prior information (domain ID) or tuning parameter, and it is based on Transformer decoder. We evaluate our Transformer Rescorer with our Joint Audio/Text training method on Librispeech dataset as well as large-scale in-house dataset and show that our model significantly outperforms over Transformer Rescorer with standard training method WER without requiring extra training parameters or computational latency. 

    
\section{Transformer Rescorer}
\label{sec:two_pass}
    The architecture of the two-pass system \citep{sainath2019two} consists of streaming 1st pass RNN-T model \citep{Graves12transduction, Shi2021emformer} and non-streaming 2nd pass Transformer Rescorer \citep{li2020parallel} is illustrated in Figure \ref{fig:two_pass}. The standard way to train the two-pass system is two-step:
    \begin{enumerate}
        \item Given paired input acoustic frames, $\bm{x} = (x_1, \cdots, x_T)$, and corresponding output text, $\bm{y} = (y_1, \cdots, x_U)$, the Encoder of RNN-T generates high-level audio embeddings, $\bm{h} = (h_1, \cdots, h_T)$. The Decoder of RNN-T takes $\bm{h}$ and $\bm{y}$ and generate initial output $\bm{\hat{y}^{1st}}$ in a streaming fashion. The RNN-T model is trained to maximize $P(\bm{y} = \bm{\hat{y}} | \bm{x})$ \citep{graves2012sequence} first. The parameters of RNN-T are then fixed.   
        \item Then, Transformer Rescorer is trained with audio embeddings $\bm{h}$ generated from the fixed encoder of RNN-T and true transcription text $\bm{y}$ to maximize $\sum_u logP(y_u| \bm{h}, y_{<u})$. \citep{vaswani2017attention}. 
    \end{enumerate} 
    
    In step 2, Transformer Rescorer is based on Transformer Decoder setup \citep{vaswani2017attention}, and the true transcription, $\bm{y}$, is the query and audio embeddings, $\bm{h}$, is the key in cross attention. 
    
    During inference, the 1st-pass RNN-T model generates n-best initial hypotheses $\hat{\bm{y}}^{1st} = (\hat{\bm{y_1}}^{1st}, \cdots, \hat{\bm{y_n}}^{1st})$ with standard beam search process. Then, Transformer Rescorer takes the n-best initial hypotheses as well as the full audio embeddings $\bm{h_{1:T}}$ from RNN-T model and computes the log probability score for each hypothesis (re-score) and the final best hypothesis $\bm{\hat{y}^{2nd}}$ is generated.
    
\section{Joint Audio/Text Training}
\label{sec:joint_method} 
    
    Our Joint Audio/Text Training method aims to leverage large amount of unpaired text-only data without requiring extra model parameter or latency. To do so, we allow the model to be trained on either 1) paired audio/text data where both modality inputs are available, or 2) unpaired text data where only text inputs are available. Our training method is also two-step as described in \ref{sec:two_pass}.  

    In step 2, for training on the unpaired text-only example, we use the estimated averaged audio embeddings, $\bm{h_{avg}}$, of paired audio/text data in the training set. This averaged audio embeddings, $\bm{h_{avg}}$, can be simply obtained by passing a 0-dummy audio sequence to the well-trained Encoder of RNN-T from the step 1. Our Transformer Rescorer takes this averaged audio embeddings instead as the keys and values in cross attention. Unlike \citep{sainath2020attention, wang2020multitask}, our approach is based on Transformer architecture and it does not need to change the original objective function, nor does it require a tuning parameter for multiple losses or any prior knowledge of inputs. Figure~\ref{fig:arch} illustrates our proposed Joint Audio/Text training method for Transformer Rescorer. Note that the inference process is the same as general rescorer as described in \ref{sec:two_pass}.

\section{Experiments}
\subsection{Data}
    \noindent \textbf{Librispeech} \hspace{0.2cm}
    We first evaluate our approach on the Librispeech English corpus \citep{panayotov2015librispeech} which is publicly available. The training data contains 960 hours of labeled speech and an additional text-only corpus containing 810M words. We apply spectrum augmentation \citep{park2019specaugment} and speed perturbation. The Librispeech unpaired text-only corpus contains 810M words and 40M samples, which is almost 27 times bigger than the paired audio/text data. We will discuss the effect of data mixing ratio of text-only data in Section \ref{sec:mixing_ratio}
    
    \noindent \textbf{Large-Scale In-house Voice Command} \hspace{0.2cm}
    \label{lab:data_inhouse}
    We also evaluate our approach on our large-scale in-house English dataset as well. Our in-house training dataset has two sources: 20K hours of publicly shared video data and 20K hours of voice assistant domain data. All videos and audios are completely de-identified. 
    We augment the training data with speed perturbation, simulated room impulse response, and background noise, resulting in 145K hours.
    Unlike Librispeech, we have 4 times smaller size of unpaired text-only corpus then paired audio-text data. The unpaired text-only data contains 1M samples of in-domain (VA) text-only data and 25M samples of general-domain text-only data. 
    Our evaluation data, VA has 44.2K de-identified short-form utterances in the voice assistant domain, collected by a third-party data vendor.

\subsection{Model}
    For the 1st pass model, we use an RNN-T model architecture which is widely used in streaming ASR \citep{graves2012sequence, graves2013speech}. The RNN-T consists of Encoder and Decoder. Our Encoder has 20 emformer layers \citep{Shi2021emformer}. We extract 80-channel filterbanks features and convert to 320-dimensional inputs with a stride of 4. The context size is 160ms for streaming restriction. The Predictor in Decoder consists of 3 LSTM layers. The Joiner in Decoder uses 5k word-piece as our targets. Our 1st-pass RNN-T model has 79M parameters.  
    
    For the 2nd pass model, Transformer Rescorer, we use 2 layers of conventional Transformer decoder \citep{vaswani2017attention} contains both the self-attention and the cross-attention. The attention dimension is 1024 and feed forward dimension is 4K, and use 8 multi-headed attention. Transformer Rescorer takes the hypothesis from the RNN-T Decoder as a query and the 1024-dimenional audio embeddings from the RNN-T Encoder as a key/value in the cross-attention. Our 2nd-pass Transformer Rescorer model has 44M parameters. Note that our Joint Audio/Text training does not require any extra model parameter. The architectures of the 1st pass RNN-T model and the 2nd pass Transformer Rescorer are illustrated in Figure \ref{fig:two_pass}. 
    
    The RNN-T and Transformer rescorer are trained in two steps as described in \ref{sec:two_pass} and \ref{sec:joint_method}. For Librispeech experiments, we trained RNN-T for 120 epochs with ADAM optimizer and a base learning rate of 0.001. For large-scale in-house experiments, we trained 15 epochs with same scheduler until full convergence. 
    
    During inference, we used the standard beam search with a beam size of 10 and generated 10-best hypotheses. As described in \ref{sec:two_pass}, Once initial decoding is done from RNN-T model, 10-best hypotheses and full context audio embeddings from RNN-T are passed to Transformer rescorer in parallel. For Librispeech experiments, we did not use any external neural language model. For large-scale in-house experiments, we used a small neural language model(2.5M parameters) for decoding with shallow fusion \citep {Kim2021lmfusion} and ILME \citep{Meng2021ILME} to compare our method with the best baseline system.
    
    We evaluated the speech engine perceived latency (SPL) for the latency analysis. The SPL measures the time from the end of the user's utterance until the ASR engine completes the result. We evaluated the SPL for three models: 1) 1st pass RNNT baseline without 2nd pass Transformer Rescorer, 2) Baseline with 2nd pass Transformer Rescorer, and 3) Baseline with 2nd pass Transformer Rescorer trained with our proposed Joint Audio/Text training method. The averaged SPLs (measured in ms) were 633.0, 636.0, 636.0, for models 1), 2) and 3), respectively. Overall, although the use of the 2nd pass model can increase SPL, our proposed Joint Audio/Text training method itself does not increase the latency at all because it does not require any model architecture changes.

    \begin{figure}[t]
        \begin{minipage}[b]{1.0\linewidth}
        \centering
        \centerline{\includegraphics[width=8cm]{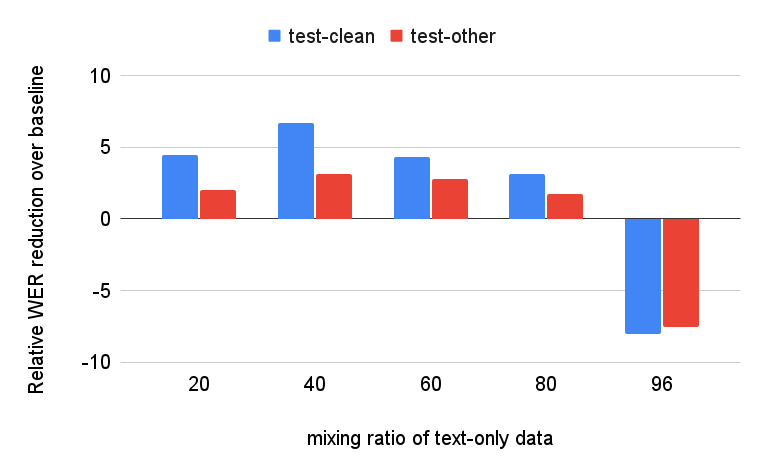}}
        \caption{WER improvement of Rescorer with different \texttt{mixing ratio} on Librispeech.}
        \label{fig:mixing}
        \end{minipage}
    \end{figure}
    
\section{Results}

\subsection{Effect of Mixing Ratio}
\label{sec:mixing_ratio}
    Similar to previous study \citep{wang2020multitask}, we observed that text-only data mixing ratio is crucial to succeed with Joint Audio/Text training for Transformer rescorer as well. Figure \ref{fig:mixing} shows the relative WER reduction of the Rescorer on Librispeech text-clean/test-other with different \texttt{mixing ratio}, defined as follows:
    \begin{align*}
        \texttt{mixing ratio} = \frac{\text{\# of text-only}}{\text{\# of (text-only + audio-text)}}
    \end{align*}
    
    Note that when we use the entire text-only data from the Librispeech provided, the mixing ratio was 96\%. The baseline in Figure \ref{fig:mixing} was 0\% mixing ratio which means that we used Rescorer trained only on paired audio/text data.  
    We observed that text-only data of 40\% \texttt{mixing ratio} performed best, and adding more than 80\% of text-only data even performed worse than the baseline Rescorer. Based on this observation, we used 40\% and 50\% \texttt{mixing ratio} for the experiments in \ref{sec:res_libri} and \ref{sec:res_inhouse}, respectively.

\subsection{Results on Librispeech}
\label{sec:res_libri}
    \begin{table}[b]
    \centering
    \resizebox{\columnwidth}{!}{
    \begin{tabular}{rrr}
    \toprule
    \textbf{Models}            & \textbf{test-clean} & \textbf{test-other} \\
    \midrule
    BS                & 3.59       & 9.10       \\
    BS + RS            & 3.27       & 8.25       \\
    BS + RS + Our Joint A/T & \textbf{3.06}       & \textbf{7.99}       \\
    \bottomrule
    \end{tabular}
    }
    \caption{Comparison of WER on Librispeech with the baseline (BS), BS with the standard rescoring model (BS + RS), and BS with RS trained by our Joint Audio/Text (BS + RS + Our Joint A/T).}
    \label{tab:res_libri}
    \end{table}
    
    Table \ref{tab:res_libri} shows WER results on Librispeech test-clean/test-other evaluation set with Baseline (BS), Baseline with Transformer Rescorer (BS + RS), and the baseline with Transformer Rescorer with our Joint audio/text training (BS + RS + Our Joint A/T). As we discussed in Section \ref{sec:mixing_ratio}, we used 40\% mixing ratio to obtain the best results. We observed that 8.9\% and 9.4\% WER relative improvements by using standard Transformer Rescorer(RS), and 14.9\% and 12.2\% WER relative improvement by using RS trained with our Joint A/T.

\subsection{Large-Scale In-house dataset}
\label{sec:res_inhouse}
    Table \ref{tab:res_inhouse} shows WER results on 44.2K in-house evaluation set with Baseline (BS), Baseline with Transformer Rescorer (BS + RS), and the baseline with the Transformer Rescorer with our Joint audio/text training (BS + RS + Our Joint A/T). As previously described in Section \ref{lab:data_inhouse}, the text-only data in test domain was only 1\% and the text-only data in general domain was only 19\% among the entire training data. In this experiment, we over-sampled in-domain text-only data to 20\% and out-of-domain text-only data to 30\%, thus we use 50\% \texttt{mixing-ratio}. We observed that 4.9\% WER relative improvements by using standard rescorer, and 7.8\% WER relative improvement by using rescorer trained with our Joint A/T. Surprisingly, our approach was still effective with our strong baseline which was trained on 145K hours. We also observed that using over-sampled duplicated in-domain text-only data is more effective rather than using unique out-of-domain text-only data.
    
    \begin{table}[t]
    \centering
    \begin{tabular}{rr}
    \toprule
    \textbf{Models}            & \textbf{VA} \\
    \midrule
    BS                & 8.32    \\
    BS + RS            & 7.91   \\
    BS + RS + Our Joint A/T   & \textbf{7.67}  \\
    \bottomrule
    \end{tabular}
    \caption{Comparison of WER on large-scale in-house dataset. VA is our in-house 20K hours of voice assistant domain data (described in Section 4.1.)}
    \label{tab:res_inhouse}
    \end{table}

\section{Conclusions}
    We have introduced Joint Audio/Text training method for Transformer Rescorer of the streaming two-pass end-to-end ASR. Unlike standard training method for Transformer Rescorer, our method can leverage unpaired text-only data and consequently improves recognition accuracy without requiring extra model parameters or computational latency. We evaluated our approach on the Librispeech dataset as well as large-scale in-house dataset and showed that Transformer Rescorer with our proposed method obtained 3\% - 7\% relative improvement in WER compared to the standard Transformer Rescorer model.

\section*{Limitations}
As with the majority of studies, this study has potential limitation. The primary limitation is that the benefit of our training approach that leverages unpaired text-only data may be diminished when paired audio/text training data is abundant and they are in same domain as test domain.





\bibliography{anthology,mybib}

\begin{thebibliography}{33}
\expandafter\ifx\csname natexlab\endcsname\relax\def\natexlab#1{#1}\fi

\bibitem[{Bahdanau et~al.(2016)Bahdanau, Chorowski, Serdyuk, Brakel, and
  Bengio}]{bahdanau2016end}
Dzmitry Bahdanau, Jan Chorowski, Dmitriy Serdyuk, Philemon Brakel, and Yoshua
  Bengio. 2016.
\newblock End-to-end attention-based large vocabulary speech recognition.
\newblock In \emph{ICASSP}. IEEE.

\bibitem[{Battenberg et~al.(2017)Battenberg, Chen, Child, Coates, Li, Liu,
  Satheesh, Sriram, and Zhu}]{battenberg2017exploring}
Eric Battenberg, Jitong Chen, Rewon Child, Adam Coates, Yashesh Gaur~Yi Li,
  Hairong Liu, Sanjeev Satheesh, Anuroop Sriram, and Zhenyao Zhu. 2017.
\newblock Exploring neural transducers for end-to-end speech recognition.
\newblock In \emph{ASRU}, pages 206--213. IEEE.

\bibitem[{Chan et~al.(2016)Chan, Jaitly, Le, and Vinyals}]{chan2016listen}
William Chan, Navdeep Jaitly, Quoc Le, and Oriol Vinyals. 2016.
\newblock Listen, attend and spell: A neural network for large vocabulary
  conversational speech recognition.
\newblock In \emph{ICASSP}. IEEE.

\bibitem[{{Chiu} et~al.(2018){Chiu}, {Sainath}, {Wu}, {Prabhavalkar}, {Nguyen},
  {Chen}, {Kannan}, {Weiss}, {Rao}, {Gonina}, {Jaitly}, {Li}, {Chorowski}, and
  {Bacchiani}}]{Chiu18}
C.~{Chiu}, T.~N. {Sainath}, Y.~{Wu}, R.~{Prabhavalkar}, P.~{Nguyen}, Z.~{Chen},
  A.~{Kannan}, R.~J. {Weiss}, K.~{Rao}, E.~{Gonina}, N.~{Jaitly}, B.~{Li},
  J.~{Chorowski}, and M.~{Bacchiani}. 2018.
\newblock {State-of-the-Art Speech Recognition with Sequence-to-Sequence
  Models}.
\newblock In \emph{ICASSP}.

\bibitem[{Chorowski et~al.(2015)Chorowski, Bahdanau, Serdyuk, Cho, and
  Bengio}]{chorowski2015attention}
Jan~K Chorowski, Dzmitry Bahdanau, Dmitriy Serdyuk, Kyunghyun Cho, and Yoshua
  Bengio. 2015.
\newblock Attention-based models for speech recognition.
\newblock In \emph{NeurIPS}.

\bibitem[{Graves(2012{\natexlab{a}})}]{Graves12transduction}
A.~Graves. 2012{\natexlab{a}}.
\newblock Sequence transduction with recurrent neural networks.
\newblock In \emph{ICML Representation Learning Workshop}.

\bibitem[{Graves(2012{\natexlab{b}})}]{graves2012sequence}
Alex Graves. 2012{\natexlab{b}}.
\newblock Sequence transduction with recurrent neural networks.
\newblock \emph{arXiv preprint arXiv:1211.3711}.

\bibitem[{Graves et~al.(2013)Graves, Mohamed, and Hinton}]{graves2013speech}
Alex Graves, Abdel-rahman Mohamed, and Geoffrey Hinton. 2013.
\newblock Speech recognition with deep recurrent neural networks.
\newblock In \emph{ICASSP}, pages 6645--6649. IEEE.

\bibitem[{Gulcehre et~al.(2015)Gulcehre, Firat, Xu, Cho, Barrault, Lin,
  Bougares, Schwenk, and Bengio}]{gulcehre2015using}
Caglar Gulcehre, Orhan Firat, Kelvin Xu, Kyunghyun Cho, Loic Barrault, Huei-Chi
  Lin, Fethi Bougares, Holger Schwenk, and Yoshua Bengio. 2015.
\newblock On using monolingual corpora in neural machine translation.
\newblock \emph{arXiv preprint arXiv:1503.03535}.

\bibitem[{He et~al.(2019)He, Sainath, Prabhavalkar, McGraw, Alvarez, Zhao,
  Rybach, Kannan, Wu, Pang et~al.}]{he2019streaming}
Yanzhang He, Tara~N Sainath, Rohit Prabhavalkar, Ian McGraw, Raziel Alvarez,
  Ding Zhao, David Rybach, Anjuli Kannan, Yonghui Wu, Ruoming Pang, et~al.
  2019.
\newblock Streaming end-to-end speech recognition for mobile devices.
\newblock In \emph{ICASSP}, pages 6381--6385. IEEE.

\bibitem[{Kannan et~al.(2018)Kannan, Wu, Nguyen, Sainath, Chen, and
  Prabhavalkar}]{kannan2018analysis}
Anjuli Kannan, Yonghui Wu, Patrick Nguyen, Tara~N Sainath, Zhijeng Chen, and
  Rohit Prabhavalkar. 2018.
\newblock An analysis of incorporating an external language model into a
  sequence-to-sequence model.
\newblock In \emph{ICASSP}, pages 1--5828. IEEE.

\bibitem[{Kim et~al.(2021{\natexlab{a}})Kim, Shangguan, Mahadeokar, Bruguier,
  Fuegen, Seltzer, and Le}]{Kim2021lmfusion}
S.~Kim, Y.~Shangguan, J.~Mahadeokar, A.~Bruguier, C.~Fuegen, M.~L. Seltzer, and
  D.~Le. 2021{\natexlab{a}}.
\newblock {Improved Neural Language Model Fusion for Streaming Recurrent Neural
  Network Transducer}.
\newblock In \emph{Proc. ICASSP}.

\bibitem[{Kim et~al.(2017)Kim, Hori, and Watanabe}]{kim2017joint}
Suyoun Kim, Takaaki Hori, and Shinji Watanabe. 2017.
\newblock Joint ctc-attention based end-to-end speech recognition using
  multi-task learning.
\newblock In \emph{ICASSP}. IEEE.

\bibitem[{Kim et~al.(2021{\natexlab{b}})Kim, Shangguan, Mahadeokar, Bruguier,
  Fuegen, Seltzer, and Le}]{kim2021improved}
Suyoun Kim, Yuan Shangguan, Jay Mahadeokar, Antoine Bruguier, Christian Fuegen,
  Michael~L Seltzer, and Duc Le. 2021{\natexlab{b}}.
\newblock Improved neural language model fusion for streaming recurrent neural
  network transducer.
\newblock In \emph{ICASSP 2021-2021 IEEE International Conference on Acoustics,
  Speech and Signal Processing (ICASSP)}, pages 7333--7337. IEEE.

\bibitem[{Li et~al.(2019)Li, Zhao, Hu, and Gong}]{li2019improving}
Jinyu Li, Rui Zhao, Hu~Hu, and Yifan Gong. 2019.
\newblock Improving rnn transducer modeling for end-to-end speech recognition.
\newblock In \emph{ASRU}, pages 114--121. IEEE.

\bibitem[{Li et~al.(2020)Li, Qin, Chiu, Pang, and He}]{li2020parallel}
Wei Li, James Qin, Chung-Cheng Chiu, Ruoming Pang, and Yanzhang He. 2020.
\newblock Parallel rescoring with transformer for streaming on-device speech
  recognition.
\newblock \emph{arXiv preprint arXiv:2008.13093}.

\bibitem[{McDermott et~al.(2019)McDermott, Sak, and Variani}]{McDermott19}
Eric McDermott, Hasim Sak, and Ehsan Variani. 2019.
\newblock A density ratio approach to language model fusion in end-to-end
  automatic speech recognition.
\newblock In \emph{ASRU}. IEEE.

\bibitem[{Meng et~al.(2021)Meng, Parthasarathy, Sun, Gaur, Kanda, Lu, Chen,
  Zhao, Li, and Gong}]{Meng2021ILME}
Z.~Meng, S.~Parthasarathy, E.~Sun, Y.~Gaur, N.~Kanda, L.~Lu, X.~Chen, R.~Zhao,
  J.~Li, and Y.~Gong. 2021.
\newblock {Internal Language Model Estimation for Domain-Adaptive End-to-End
  Speech Recognition}.
\newblock In \emph{Proc. SLT}.

\bibitem[{Panayotov et~al.(2015)Panayotov, Chen, Povey, and
  Khudanpur}]{panayotov2015librispeech}
Vassil Panayotov, Guoguo Chen, Daniel Povey, and Sanjeev Khudanpur. 2015.
\newblock {LibriSpeech: an ASR corpus based on public domain audio books}.
\newblock In \emph{ICASSP}.

\bibitem[{Park et~al.(2019)Park, Chan, Zhang, Chiu, Zoph, Cubuk, and
  Le}]{park2019specaugment}
Daniel~S Park, William Chan, Yu~Zhang, Chung-Cheng Chiu, Barret Zoph, Ekin~D
  Cubuk, and Quoc~V Le. 2019.
\newblock Specaugment: A simple data augmentation method for automatic speech
  recognition.
\newblock \emph{arXiv preprint arXiv:1904.08779}.

\bibitem[{Prabhavalkar et~al.(2017)Prabhavalkar, Rao, Sainath, Li, Johnson, and
  Jaitly}]{Prabhavalkar17}
R.~Prabhavalkar, K.~Rao, T.~Sainath, B.~Li, L.~Johnson, and N.~Jaitly. 2017.
\newblock {A Comparison of Sequence-to-Sequence Models for Speech Recognition}.
\newblock In \emph{Interspeech}, pages 939--943.

\bibitem[{Sainath et~al.(2019)Sainath, Pang, Rybach, He, Prabhavalkar, Li,
  Visontai, Liang, Strohman, Wu et~al.}]{sainath2019two}
Tara~N Sainath, Ruoming Pang, David Rybach, Yanzhang He, Rohit Prabhavalkar,
  Wei Li, Mirk{\'o} Visontai, Qiao Liang, Trevor Strohman, Yonghui Wu, et~al.
  2019.
\newblock Two-pass end-to-end speech recognition.
\newblock \emph{arXiv preprint arXiv:1908.10992}.

\bibitem[{Sainath et~al.(2020)Sainath, Pang, Weiss, He, Chiu, and
  Strohman}]{sainath2020attention}
Tara~N Sainath, Ruoming Pang, Ron~J Weiss, Yanzhang He, Chung-cheng Chiu, and
  Trevor Strohman. 2020.
\newblock An attention-based joint acoustic and text on-device end-to-end
  model.
\newblock In \emph{ICASSP 2020-2020 IEEE International Conference on Acoustics,
  Speech and Signal Processing (ICASSP)}, pages 7039--7043. IEEE.

\bibitem[{Shan et~al.(2019)Shan, Weng, Wang, Su, Luo, Yu, and
  Xie}]{shan2019component}
Changhao Shan, Chao Weng, Guangsen Wang, Dan Su, Min Luo, Dong Yu, and Lei Xie.
  2019.
\newblock Component fusion: Learning replaceable language model component for
  end-to-end speech recognition system.
\newblock In \emph{ICASSP}, pages 5361--5635. IEEE.

\bibitem[{Shi et~al.(2021)Shi, Wang, Wu, Yeh, Chan, Zhang, Le, and
  Seltzer}]{Shi2021emformer}
Y.~Shi, Y.~Wang, C.~Wu, C.~Yeh, J.~Chan, F.~Zhang, D.~Le, and M.~L. Seltzer.
  2021.
\newblock {Emformer: Efficient Memory Transformer Based Acoustic Model For Low
  Latency Streaming Speech Recognition}.
\newblock In \emph{Proc. ICASSP}.

\bibitem[{Sriram et~al.(2017)Sriram, Jun, Satheesh, and
  Coates}]{sriram2017cold}
Anuroop Sriram, Heewoo Jun, Sanjeev Satheesh, and Adam Coates. 2017.
\newblock Cold fusion: Training seq2seq models together with language models.
\newblock \emph{arXiv preprint arXiv:1708.06426}.

\bibitem[{Tang et~al.(2022)Tang, Gong, Dong, Wang, Hsu, Gu, Baevski, Li,
  Mohamed, Auli et~al.}]{tang2022unified}
Yun Tang, Hongyu Gong, Ning Dong, Changhan Wang, Wei-Ning Hsu, Jiatao Gu,
  Alexei Baevski, Xian Li, Abdelrahman Mohamed, Michael Auli, et~al. 2022.
\newblock Unified speech-text pre-training for speech translation and
  recognition.
\newblock \emph{arXiv preprint arXiv:2204.05409}.

\bibitem[{Toshniwal et~al.(2018)Toshniwal, Kannan, Chiu, Wu, Sainath, and
  Livescu}]{toshniwal2018comparison}
Shubham Toshniwal, Anjuli Kannan, Chung-Cheng Chiu, Yonghui Wu, Tara~N Sainath,
  and Karen Livescu. 2018.
\newblock A comparison of techniques for language model integration in
  encoder-decoder speech recognition.
\newblock In \emph{SLT}, pages 369--375. IEEE.

\bibitem[{Variani et~al.(2020)Variani, Rybach, Allauzen, and
  Riley}]{variani2020hybrid}
Ehsan Variani, David Rybach, Cyril Allauzen, and Michael Riley. 2020.
\newblock Hybrid autoregressive transducer (hat).
\newblock In \emph{ICASSP}, pages 6139--6143. IEEE.

\bibitem[{Vaswani et~al.(2017)Vaswani, Shazeer, Parmar, Uszkoreit, Jones,
  Gomez, Kaiser, and Polosukhin}]{vaswani2017attention}
Ashish Vaswani, Noam Shazeer, Niki Parmar, Jakob Uszkoreit, Llion Jones,
  Aidan~N Gomez, {\L}ukasz Kaiser, and Illia Polosukhin. 2017.
\newblock Attention is all you need.
\newblock In \emph{NeurIPS}.

\bibitem[{Wang et~al.(2020)Wang, Sainath, and Weiss}]{wang2020multitask}
Peidong Wang, Tara~N Sainath, and Ron~J Weiss. 2020.
\newblock Multitask training with text data for end-to-end speech recognition.
\newblock \emph{arXiv preprint arXiv:2010.14318}.

\bibitem[{Weinstein et~al.(2020)Weinstein, Apfel, Ghodsi, Cabrera, and
  Liu}]{weinstein2020stateless}
Eugene Weinstein, James Apfel, Mohammadreza Ghodsi, Rodrigo Cabrera, and
  Xiaofeng Liu. 2020.
\newblock Rnn-transducer with stateless prediction network.
\newblock In \emph{ICASSP}, pages 7049--7053.

\bibitem[{Xu et~al.(2022)Xu, Gu, Kolehmainen, Khan, Gandhe, Rastrow, Stolcke,
  and Bulyko}]{xu2022rescorebert}
Liyan Xu, Yile Gu, Jari Kolehmainen, Haidar Khan, Ankur Gandhe, Ariya Rastrow,
  Andreas Stolcke, and Ivan Bulyko. 2022.
\newblock Rescorebert: Discriminative speech recognition rescoring with bert.
\newblock In \emph{ICASSP 2022-2022 IEEE International Conference on Acoustics,
  Speech and Signal Processing (ICASSP)}, pages 6117--6121. IEEE.

\end{thebibliography}
\bibliographystyle{acl_natbib}


\end{document}